\documentclass[11pt]{article}

\PassOptionsToPackage{hyphens}{url}
\usepackage{acl}

\usepackage{tempora}
\usepackage{latexsym}

\usepackage[T1,T2A]{fontenc}
\usepackage[utf8]{inputenc}
\usepackage[russian,english]{babel}

\usepackage{microtype}
\usepackage{inconsolata}
\let\benchttfamily\ttfamily
\renewcommand{\ttfamily}{\fontencoding{T1}\benchttfamily}
\usepackage{graphicx}

\usepackage{booktabs}
\usepackage{multirow}
\usepackage{array}
\usepackage{amsmath,amssymb}
\usepackage{xcolor}
\usepackage{tikz}
\usetikzlibrary{arrows.meta,positioning,fit,backgrounds,calc}

\newcommand{\bench}{WebPageBench}
\newcommand{\ems}{EMS}

\title{\bench{}: Event-Level Verification and Controlled UI-Variant Generation for Web Agents}

\author{
  Anton Emelyanov\textsuperscript{1} \quad Maria Tikhonova\textsuperscript{1,2} \quad Zaven Martirosian\textsuperscript{1,3} \quad
  \\ \textbf{Sergei Averkiev\textsuperscript{1} \quad Alena Fenogenova\textsuperscript{1,2}} \\
  \textsuperscript{1}DAIMLD \qquad \textsuperscript{2}HSE University \qquad \textsuperscript{3}NUST MISIS \\
  \texttt{login-const@mail.ru} 
}

\begin{document}
\maketitle

\begin{abstract}
We present \bench{}, an open framework for evaluating web agents in which every task is verified from the interface's own event log. Six instrumented mock sites with brand identifiers removed (a marketplace, a bookstore, a grocery service, rail ticketing, hotel search and a document cabinet) emit typed events with parameters as a user or an agent acts. A task declares the events it requires, and success is decided by matching them, with no judge model and no scraping of rendered pages. The same instrumentation supports controlled UI variation: one configuration switch re-renders a task through a different implementation of a single control while the prompt and the success conditions stay completely identical, so sensitivity to interface form can be measured under a fixed task specification. The \bench{} release consists of three components: 152 tasks, divided into 65 canonical scenarios and 87 control variants across light/dark UI-modes; a common runner evaluated with six browser/DOM harness configurations and five screenshot-only GUI-agent families; and a public leaderboard of 24 model--harness pairs. On the public 152-task leaderboard the gap between what agents declare finished and what the log confirms reaches 41 points (one configuration declares every task finished and satisfies the conditions on 59\%). %
\end{abstract}

\section{Introduction}
\label{sec:intro}

\begin{figure*}[t]
\centering
\includegraphics[width=\textwidth]{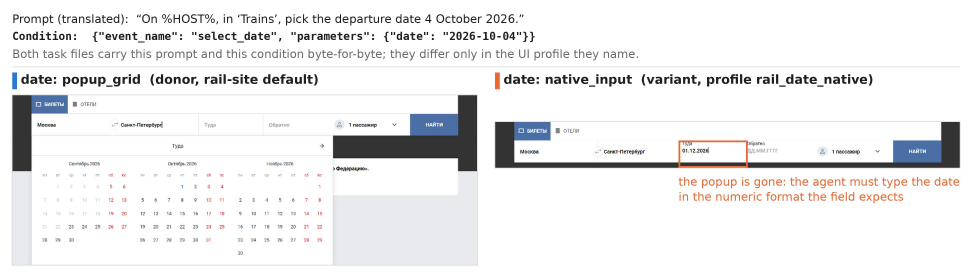}
\caption{One task, one condition, two interfaces. The prompt (translated) and the single success condition shown above the screenshots are identical in both task files; the released variant differs from its donor only in the profile it names (\texttt{rail\_date\_native}). On the left the departure date is picked from the default popup grid, where the month and the day are on screen and one click settles it. On the right the same date goes into a native \texttt{<input type="date">}, so the agent has to render ``4 October 2026'' from the prompt in the numeric format the field expects (the screenshot shows the field with an arbitrary date typed in). The figure illustrates the variant mechanism only; the leaderboard submissions do not separate donors from variants, so no per-variant result is reported.}
\label{fig:teaser}
\end{figure*}

Benchmarks for web agents trade verifiability against scale. Offline trajectory matching \citep{deng2023mind2web} is reproducible but scores the path instead of the outcome. Live-web benchmarks \citep{he2024webvoyager,yoran2024assistantbench,trabucco2025insta} reach hundreds of sites and pay for it with a judge model, drifting content and anti-bot defenses. Self-hosted environments \citep{zhou2023webarena,boisvert2024workarena,garg2025real,xu2024theagentcompany} check application state directly, but only the state that remains when the episode ends, so the steps an agent took and the state it left behind are indistinguishable.

Existing benchmarks additionally fail to isolate agent sensitivity to the implementation of individual UI controls. Figure~\ref{fig:teaser} shows what such a comparison requires: the same prompt and the same success condition rendered once through a popup calendar and once through a native date field. An aggregate score over ordinary benchmarks cannot tell a failure on the second interface from a misunderstood task.


\bench{} addresses both. It is intended for researchers and engineers who develop web agents and need to tell model failures from harness and interface failures, and to reproduce those failures locally. The environment is a set of mock sites we control and instrument: every action a user or an agent performs writes a typed event with parameters to a backend, and a task's success condition is a list of such events. 
Because the interface is configurable, one switch produces a variant of a task in which only the control changes; the prompt and the conditions are unchanged, so a difference in success rate reflects the interface realisation. Recording the agent's completion signal separately from the verifier already gives a result: in the runs we report, agents declare success far more often than the log confirms it, and the size of the gap depends on the harness. Interfaces, prompts and data are Russian; the verification contract reads typed events and never the interface text, so it does not depend on the language.

The main contribution of this work is \bench{}, an evaluation system for web agents built around one methodology: the interface is instrumented, so success is decided from the typed events it emits, and the same instrumentation makes the interface itself a controlled variable. The system consists of three components, each of which follows from that principle.
\begin{itemize}\itemsep2pt
  \item \textbf{The event-level verification contract} (\S\ref{sec:events}, \S\ref{sec:contract}) defines what counts as success: 19 event types with typed parameters, per-task conditions with exact, wildcard, comparison and last-state matching, and a primary metric, the Event-Match Score (\ems{}), that needs no judge model. The agent's own completion signal is recorded separately, so over-claiming becomes measurable.
  \item \textbf{The generator of controlled UI variants} (\S\ref{sec:variants}) makes the interface a variable of the experiment: nine UI configuration keys (eight controls and the colour theme) with 33 implementations produce 87 variant tasks from 25 donors, 82 of them changing exactly one control, while the prompt and the conditions stay fixed.
  \item \textbf{The runner and leaderboard} (\S\ref{sec:demo}) apply the contract uniformly to both action spaces: six evaluated browser/DOM harness configurations and five screenshot-only GUI-agent families run under one verifier, so a comparison can hold the model, the harness or the interface fixed; the release adds a reproducible local deployment, automatic repair of time-dependent tasks, and a public leaderboard of 24 model--harness pairs on the 152-task suite.
\end{itemize}

\section{Related Work}
\label{sec:related}

Table~\ref{tab:related} groups web-agent benchmarks by what decides success, its first column. Reference matching never inspects the environment. Judge models scale to the open web but inherit the judge's failure modes, and their reliability outside English has, to our knowledge, not been established. Hybrid designs check the state of the application but fall back on a model for information-seeking tasks or images. Deterministic designs decide success programmatically: a reward over product attributes \citep{yao2022webshop}, backend API checks \citep{boisvert2024workarena}, replay against a state machine \citep{wu2026autowebworld}, final-state comparison \citep{yuan2026webforge}. \bench{} belongs to this last group and differs within it: conditions are matched against the typed events the interface emitted during the run, including intermediate steps and their parameters, instead of the state left at the end. Other work automates task construction \citep{pahuja2025explorer,xie2025agentsynth,chen2025graph2eval,fan2026webfactory,huang2026gta}; we author tasks by hand and automate the variants.

The two right-hand columns of Table~\ref{tab:related} separate \bench{} from the rest of its group. Among the systems surveyed, none renders one task through alternative implementations of a control while holding prompt and verifier fixed; the closest is WorkArena++, which samples one of ten fictitious company brands per episode, a repaint of the same interface with the controls unchanged. Where several agents can be run against a benchmark, they usually arrive through BrowserGym and AgentLab \citep{chezelles2025browsergym}, an ecosystem that wraps benchmarks behind a shared agent interface. That work is complementary: our environment could be exposed through it, and the frameworks and GUI models our runner drives are not among the agents it ships. Compared with a WebArena-style environment plus logging, four things differ: tasks are written against the event contract, the verdict is inspectable per condition, the task specification stays fixed across UI realisations, and one runner serves both action spaces.

\section{Framework Methodology}
\label{sec:benchmark}

Figure~\ref{fig:architecture} shows one run end to end. A task file supplies the prompt and the success conditions, the domain configuration supplies the control implementations, and the two are opened as an isolated track on a mock site. The agent acts on the pages that track serves. The site writes typed events to the track log, and \texttt{check()} decides success by matching the conditions against that log alone.

\subsection{Mock sites and anonymization}
\label{sec:sites}
The canonical scenarios were written by hand, not generated. The contributors who built each mock site worked out the frequent purchases and the other ordinary actions on that site, and wrote the prompt and the success conditions for every scenario. Variants (\S\ref{sec:variants}) copy those scenarios and change a control; they are not new scenarios.

Tasks live under one neutral hub with six tabs, each a mock site covering an everyday online activity: marketplace, books and audiobooks, grocery delivery, rail travel, hotel search, and a file document cabinet. The sites are written as one front end over a back end that serves per-task configuration and stores events.

Anonymization goes beyond the logos. The mock sites are based on real, widely used platforms and are anonymized: route and state names are neutral (\texttt{bench\_\allowbreak catalog\_\allowbreak main}, \texttt{bench\_\allowbreak hotel\_\allowbreak search}), so neither the URL nor the page structure gives a model any navigational or branding cue to a well-known platform whose flow it may have memorized. The book catalog is additionally rewritten by a deterministic script (seed 42): 226 items, 75 author names, 229 titles, and 86 series are replaced with synthetic Russian names, generated together with their grammatical case forms so that prompts naming an author remain well-formed.
Book covers are dropped in that same pass. By contrast, the marketplace and grocery catalogs are not textually rewritten:
their accompanying text is left intact, while the card photographs are replaced with openly licensed files stored locally.
After image pruning the marketplace holds 42 products and the grocery service 277 products across 117 category files. The release contains 590 active card-image paths: 277 grocery, 41 marketplace and 272 hotel images, linked to 444 distinct source URLs. The attribution workbook maps every local path to its source URL and license\footnote{The license distribution is 419 CC~BY~2.0, 145 CC~BY-SA~2.0, 8 CC~BY-SA~4.0, 2 CC~BY-SA~3.0, 12 CC0~1.0 and 4 Public Domain images; no NC or ND license is included.}; 543 rows also include an Openverse image-information page. The workbook contains neither similarity scores nor automatic/manual decision flags, so we do not report a 0.5 threshold or an automatic-vs.-manual split. The verifier itself does not inspect image content: conditions match typed events rather than pixels.

\subsection{Events and UI configuration keys}
\label{sec:events}

A \textbf{track} is one isolated run of one task: it has its own URL, application state and event log, so parallel runs never see each other. An \textbf{event} is a typed record the interface writes to the track when something happens: a name and a dictionary of parameters. \texttt{basket\_add} carries the item identifier and quantity, \texttt{select\_tariff} the tariff name, \texttt{bench\_hotel\_select\_city} the city identifier, display name and country. Events are emitted by the pages themselves and never reconstructed from the DOM, so they record what the application registered, independently of how the page looked. Low-level activity (\texttt{click}, \texttt{keypress}, \texttt{scroll}) is logged too, but conditions refer only to semantic events: the suite uses 19 such event types, and the 152 released tasks declare 521 conditions over them, 3.4 per task on average.

A \textbf{UI configuration key} names a control whose implementation is selected by configuration instead of being fixed in the markup; eight keys are controls and the ninth is the colour theme. Picking a date is one interaction; the hotel site can render it as a split popup calendar (the default), an always-visible inline calendar, a typed DD.MM.YYYY string, or a single-field range picker, and the rail site as a grid popup (its default), a native \texttt{<input type="date">} or the same typed string, so the \texttt{date} key has six implementations. Choosing a city is an autocomplete or a native \texttt{select}; a guest counter is a per-room popup, an inline stepper, a compact \texttt{select} or a row of pill buttons. The nine keys have 33 implementations in total (Appendix~\ref{app:variants}). A control emits the same event whichever implementation is mounted (Figure~\ref{fig:event}), so the condition that checks it stays valid across implementations.

\begin{figure}[t]
\centering
\footnotesize
\begin{tabular}{@{}p{0.95\linewidth}@{}}
\toprule
\textbf{Condition declared by the task} \\
\begin{minipage}{\linewidth}
\begin{verbatim}
{"event_name": "bench_hotel_select_guests",
 "parameters": {"roomsCount": 1,
                "guestsCount": 2}}
\end{verbatim}
\end{minipage} \\
\midrule
\textbf{Event written by the interface} \\
\begin{minipage}{\linewidth}
\begin{verbatim}
{"type": "bench_hotel_select_guests",
 "roomsCount": 1, "guestsCount": 2,
 "widget": "pill_buttons"}
\end{verbatim}
\end{minipage} \\
\bottomrule
\end{tabular}
\caption{A condition and the event that satisfies it. All four implementations of the guest counter (per-room popup, inline stepper, compact \texttt{select}, pill buttons) emit the same event with the same parameters, so the condition holds for all of them. The \texttt{widget} field is recorded for analysis and never enters the verdict.}
\label{fig:event}
\end{figure}

\begin{figure}[t]
\centering
\resizebox{0.62\linewidth}{!}{%
\begin{tikzpicture}[
  node distance=3.2mm,
  box/.style={draw, rounded corners=1.5pt, align=center, font=\scriptsize, inner sep=3pt, minimum height=6.5mm, text width=34mm},
  dark/.style={box, fill=black!6},
  arr/.style={-{Latex[length=1.6mm]}, thin}
]
\node[dark] (task) {Task JSON\\\textit{prompt + conditions}};
\node[dark, right=6mm of task] (cfg) {Domain config\\\textit{UI profile}};
\coordinate (mid) at ($(task.south)!0.5!(cfg.south)$);
\node[box, below=of mid, text width=70mm] (track) {Track \textit{(isolated run)}};
\node[box, below=of track] (site) {Mock site};
\node[box, below=of site] (log) {Event log \textit{(typed, parameterized)}};
\node[box, below=of log] (check) {\texttt{check()}: conditions\\vs.\ events};
\node[box, below=of check] (score) {\ems{} $+$ agent\\self-report};
\node[box, left=6mm of site, text width=20mm] (agent) {Agent harness};
\draw[arr] (task.south) -- (track.north -| task.south);
\draw[arr] (cfg.south) -- (track.north -| cfg.south);
\draw[arr] (track) -- (site);
\draw[arr] (site) -- (log);
\draw[arr] (log) -- (check);
\draw[arr] (check) -- (score);
\draw[{Latex[length=1.6mm]}-{Latex[length=1.6mm]}, thin] (agent) -- (site);
\end{tikzpicture}%
}
\caption{One run. The task supplies the prompt and the conditions, the domain configuration supplies the control implementations, and the event log is the only evidence used to decide success.}
\label{fig:architecture}
\end{figure}
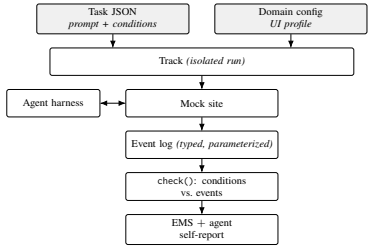

Figure~\ref{fig:architecture} shows how the two meet: a task JSON and the domain configuration are instantiated as a track, the agent acts against the served pages, and the client library checks the log against the conditions.

\subsection{The verification contract}
\label{sec:contract}

A condition is a predicate over the events emitted during a track: it names an event and the parameters that must match, and it holds if some event of that name carries the required parameters. Conditions are evaluated independently over the track log. Strings match exactly or through a \texttt{*} wildcard, numeric fields may carry a comparison operator, and conditions can be grouped so that satisfying any member satisfies the group. The verifier does not require a particular ordering unless the ordering is encoded indirectly in event parameters or in the application's own flow; what a task pins down is which steps happened and with which values. Quantities are the exception to ``any event'': every basket, seat and guest-count event carries the resulting quantity, and a condition on that quantity is compared with the last such event for the item: adding two units and then a third fails a condition asking for two. A task can override this per condition with \texttt{match: any} when the intermediate action is what is being tested.

The primary metric, the binary task-level \textbf{Event-Match Score (\ems{})}, is 1 for a task when every condition is satisfied and 0 otherwise; we report its mean over tasks, and the per-condition verdicts are kept as diagnostic (\S\ref{sec:demo}): in the rail task of Appendix~\ref{app:examples}, a wrong tariff satisfies six of seven conditions and scores 0. No model is consulted. Because the contract reads typed events and never the interface text, it does not depend on the language of the interface, whereas a judge model's accuracy does.

The evaluation covers 11 \textbf{Harness} identifiers (Table~\ref{tab:harnesses}): six browser integrations and five screenshot-only GUI adapters. \emph{DOM} is an indexed element tree the agent acts on by element handle; \emph{text} is extracted page text without that tree. Browser-use, Ouroboros-cut and OpenManus receive DOM plus a screenshot, except the DeepSeek pairs, which receive DOM text only; the two full Ouroboros runtimes receive text plus a screenshot; OpenHands receives DOM text only. The GUI adapters receive screenshots only and return pointer coordinates through our Playwright driver. Several integrations reuse \texttt{browser-use} as the execution layer and change only prompting, planning or memory. Both action spaces run over one suite with one verifier. The public leaderboard reports 24 evaluated model--harness pairs.

\begin{table}[t]
\centering
\scriptsize
\setlength{\tabcolsep}{2pt}
\begin{tabular}{@{}lll@{}}
\toprule
\textbf{Harness} & \textbf{Obs.} & \textbf{Notes} \\
\midrule
\texttt{browser-use} & \shortstack[l]{DOM+\\screenshot}$^\ddagger$ & baseline executor \\
\texttt{ouroboros-cut} & \shortstack[l]{DOM+\\screenshot} & Ouroboros prompts, browser-use \\
\texttt{ouroboros-full-isolated} & \shortstack[l]{text+\\screenshot} & Ouroboros, fresh memory \\
\texttt{ouroboros-full-evolving} & \shortstack[l]{text+\\screenshot} & Ouroboros, shared memory$^\dagger$ \\
\texttt{openmanus} & \shortstack[l]{DOM+\\screenshot}$^\ddagger$ & OpenManus prompts, browser-use \\
\texttt{openhands} & DOM text & OpenHands SDK, browser tools$^\dagger$ \\
\midrule
\texttt{qwen3-vl} & screenshot & Qwen3.8-27B \\
\texttt{uitars} & screenshot & UI-TARS-1.5-7B \\
\texttt{opencua} & screenshot & OpenCUA-32B / 72B \\
\texttt{evocua} & screenshot & EvoCUA-32B S1 / S2 \\
\texttt{fara} & screenshot & Fara-1.5 9B / 27B \\
\bottomrule
\end{tabular}
\caption{Harnesses evaluated in this snapshot; all are selected by the \texttt{AGENT\_HARNESS} switch: browser-use \citep{browseruse2024}, OpenManus \citep{openmanus2025}, OpenHands \citep{wang2025openhands}, Qwen3-VL \citep{bai2025qwen3vl}, UI-TARS \citep{uitars2025}, OpenCUA \citep{wang2025opencua}, EvoCUA \citep{xue2026evocua} and Fara-1.5 \citep{fara152026}; Ouroboros is an in-house agent runtime. \emph{DOM} is an indexed element tree the agent can act on by element handle; \emph{text} is extracted page text without that tree. A screenshot, when listed, is sent in addition. $^\ddagger$The DeepSeek pairs receive DOM text only, without screenshots. $^\dagger$Runs on a single worker; for the evolving runtime because tasks share one memory.}
\label{tab:harnesses}
\end{table}

Two further signals are recorded. \textbf{Completion} is the harness's own claim that it finished, for example, \texttt{browser-use's} \texttt{done} action; a run that ends in a harness error counts as not completed. Where the two disagree, we distinguish an \textbf{over-claim} (completed, EMS 0) from an \textbf{under-claim} (EMS 1, not completed). We call the difference between Completion and EMS the \emph{completion--verification gap}; \S\ref{sec:results} shows that it varies widely with the harness. Steps, time, tokens and cost are recorded as well. The published submission stores Completion and EMS as rates, together with steps, duration and cost. It does not store over-claim, under-claim or harness-error counts. Done\&pass, counted from the run logs, recovers the first two: Completion minus Done\&pass is the over-claim rate, and \ems{} minus Done\&pass is the under-claim rate.

Time-dependent tasks are repaired automatically. Booking interfaces disable past dates. Before a track is created, all date conditions are shifted forward by one shared offset so the earliest falls on or after today, and the dates in the Russian prompt (dotted, long-form and range spellings) are rewritten to match, so the suite stays runnable without editing task files.

\subsection{Generating controlled UI variants}
\label{sec:variants}

Operationally, a task is a \textbf{variant} when its file carries a variant profile and \textbf{canonical} otherwise; every canonical task in the release was authored by hand. A variant is a materialised copy of a canonical task (its \emph{donor}) with a \emph{profile} applied, a named assignment of UI configuration keys such as \texttt{hotels\_date\_inline}. The prompt and the conditions are copied, so the two share one task specification and differ only in the implementation of the control and what it brings with it: DOM structure, action count, layout, screenshot coordinates. The released suite has 87 variants from 25 donors through 20 profiles. Nineteen profiles set a single key, so 82 variants change exactly one control; one profile for the document cabinet sets three keys (layout, year selector, button style) that share a screen, giving five three-key variants. The dark theme is never combined with a control change. Table~\ref{tab:domains} gives the per-domain counts, and Table~\ref{tab:examples} in Appendix~\ref{app:examples} shows one donor per domain with its variants. Every task also carries \textbf{interaction classes} from a 19-class taxonomy (navigation, basket, date, counter, payment, and so on), one of them designated as primary; 13 classes occur in the released tasks and 9 as primary (Appendix~\ref{app:taxonomy}).

For a harness $h$ and a key $w$, let $D_w$ be the donor tasks that have a variant on $w$ and $V_w$ those variants. The sensitivity gap is
\[
\Delta_{h,w} \;=\; \mathrm{EMS}_h(D_w) - \mathrm{EMS}_h(V_w),
\]
the change in success on the same task specifications when the implementation of $w$ changes. A large $|\Delta_{h,w}|$ shows that performance is sensitive to the realisation of that control; whether the agent misunderstood the task or could not operate the control is a hypothesis to examine on the paired traces; $\Delta$ alone does not decide it. Restricting the baseline to $D_w$ would keep the task mix identical. The published submissions do not split donors from variants, so this snapshot does not report $\Delta_{h,w}$.

\begin{table}[t]
\centering
\small
\resizebox{\linewidth}{!}{%
\begin{tabular}{@{}llrrr@{}}
\toprule
\textbf{Tab} & \textbf{Activity} & \textbf{Canon.} & \textbf{Var.} & \textbf{Total} \\
\midrule
\foreignlanguage{russian}{Маркет} & marketplace & 18 & 14 & 32 \\
\foreignlanguage{russian}{Книги} & books, audiobooks & 11 & 10 & 21 \\
\foreignlanguage{russian}{Файлы} & document cabinet & 11 & 7 & 18 \\
\foreignlanguage{russian}{Поезда} & rail travel & 9 & 17 & 26 \\
\foreignlanguage{russian}{Продукты} & grocery delivery & 8 & 2 & 10 \\
\foreignlanguage{russian}{Отели} & hotel search & 8 & 37 & 45 \\
\midrule
\textbf{Total} & & \textbf{65} & \textbf{87} & \textbf{152} \\
\bottomrule
\end{tabular}%
}
\caption{Tasks per domain. Variant counts reflect the profiles available in a domain and the donors eligible for them. For example, the hotel domain combines date, city and guest-counter controls and yields 37 variants from 8 canonical tasks.} 
\label{tab:domains}
\end{table}



\section{System Demonstration}
\label{sec:demo}

\paragraph{Running the benchmark.} The stack is one Docker composition: backend with event store, front end, evaluation container. A run is one command; the harness and the model are chosen by two environment variables, and the runner creates a track per task, hands its URL to the agent, waits for the agent to stop, and calls \texttt{check()}. Evaluation is a \texttt{pytest} suite built on DeepEval \citep{deepeval2024}, parallel over workers except for the two harnesses marked in Table~\ref{tab:harnesses}. On the public leaderboard a 152-task run with \texttt{gemini-3.8-flash} and \texttt{openmanus} averages 7.8 steps and 221~s of harness time per task (median 131~s; 16.3 tasks/hour/worker at that mean; \$17.88 for the run). The submission does not record a separate cold-start time. OpenHands runs on one worker. Local GUI checkpoints in this snapshot do not report USD cost.

\paragraph{Inspecting a run.} A user picks a task, a model, a harness and a UI profile, starts a track and follows the browser episode. When the agent stops, the track page lists every event with its parameters next to the conditions and marks which matched, so the visitor sees the agent declare the task finished and, on the same page, the conditions no event satisfied. The visitor then re-runs the identical task with another implementation of one control, compares the two traces, and opens the leaderboard slice for that harness, class or key.


\begin{table*}[t]
\centering
\footnotesize
\setlength{\tabcolsep}{4pt}
\setlength{\aboverulesep}{1pt}
\setlength{\belowrulesep}{1pt}
\renewcommand{\arraystretch}{0.95}
\resizebox{0.85\textwidth}{!}{%
\begin{tabular}{@{}llrrrrrr@{}}
\toprule
\textbf{Model} & \textbf{Harness} & \textbf{\ems{}} & \textbf{Compl.} & \textbf{Done\&pass} & \textbf{Steps} & \textbf{Dur.} & \textbf{Cost \$} \\
\midrule
\multirow{3}{*}{gemini-3.8-flash}
  & \texttt{openmanus} & \textbf{0.822} & 1.000 & 0.822 & 7.8 & 221 & \textbf{17.88} \\
  & \texttt{browser-use} & 0.711 & 0.993 & 0.711 & 9.5 & 220 & 23.65 \\
  & \texttt{openhands} & 0.572 & 0.461 & 0.454 & 131.7 & \textbf{132} & 53.12 \\
\midrule
\multirow{5}{*}{gpt-5.6-luna}
  & \texttt{ouro-full-iso} & \textbf{0.796} & 0.868 & 0.796 & 27.8 & 177 & 6.69 \\
  & \texttt{browser-use} & 0.704 & 0.993 & 0.704 & 10.8 & 356 & 3.23 \\
  & \texttt{ouro-full-env} & 0.697 & 0.717 & 0.678 & 13.1 & 297 & 3.91 \\
  & \texttt{openhands} & 0.625 & 0.638 & 0.605 & 18.1 & \textbf{158} & 23.19 \\
  & \texttt{openmanus} & 0.592 & 1.000 & 0.592 & 10.8 & 296 & \textbf{3.04} \\
\midrule
\multirow{3}{*}{deepseek-v4.1-flash}
  & \texttt{browser-use} & \textbf{0.789} & 0.993 & 0.783 & 14.2 & 467 & \textbf{8.28} \\
  & \texttt{openmanus} & 0.711 & 0.974 & 0.704 & 14.7 & 472 & 9.03 \\
  & \texttt{openhands} & 0.454 & 0.342 & 0.336 & 23.5 & \textbf{148} & 39.66 \\
\midrule
\multirow{4}{*}{qwen3.8-27b}
  & \texttt{openmanus} & \textbf{0.763} & 1.000 & 0.763 & 9.9 & 84 & --- \\
  & \texttt{ouro-cut} & 0.750 & 1.000 & 0.750 & 10.4 & 83 & --- \\
  & \texttt{browser-use} & 0.737 & 1.000 & 0.737 & 10.8 & 91 & --- \\
  & \texttt{qwen3-vl} & 0.480 & 0.493 & 0.375 & 17.0 & \textbf{74} & --- \\
\midrule
fara-1.5-9b & \texttt{fara} & 0.638 & 0.572 & 0.493 & 16.4 & 44 & --- \\
fara-1.5-27b & \texttt{fara} & 0.625 & 0.625 & 0.553 & 16.5 & 72 & --- \\
glm-5.2 & \texttt{openhands} & 0.500 & 0.428 & 0.408 & 22.7 & 90 & 76.08 \\
opencua-72b & \texttt{opencua} & 0.421 & 0.724 & 0.421 & 13.2 & 437 & --- \\
minimax-m2.7 & \texttt{openhands} & 0.368 & 0.257 & 0.230 & 21.5 & 103 & 24.96 \\
opencua-32b & \texttt{opencua} & 0.336 & 0.559 & 0.336 & 15.3 & 271 & --- \\
uitars-1.5-7b & \texttt{uitars} & 0.336 & 0.382 & 0.289 & 19.1 & 42 & --- \\
evocua-32b-s1 & \texttt{evocua} & 0.092 & 0.112 & 0.079 & 23.5 & 259 & --- \\
evocua-32b-s2 & \texttt{evocua} & 0.079 & 0.191 & 0.072 & 22.5 & 143 & --- \\
\bottomrule
\end{tabular}%
}
\caption{Public leaderboard submissions (24 model--harness pairs), grouped by model. Models are ordered by their best \ems{}, rows within a model by \ems{}. Where a model has several harnesses, bold marks the highest \ems{}, the shortest duration and the lowest cost in that group; single-harness rows are not bold. \texttt{ouro-full-iso} is \texttt{ouroboros-full-isolated}, \texttt{ouro-full-env} is \texttt{ouroboros-full-evolving}, \texttt{ouro-cut} is \texttt{ouroboros-cut}. \emph{\ems{}} is \texttt{passed\_tasks}/152. \emph{Compl.} is \texttt{agent\_completion\_rate}. \emph{Done\&pass} is the share of tasks the agent reported done and whose conditions all passed, counted from the run logs; it is the same count the public leaderboard shows. \emph{Steps} and \emph{Dur.} are per-task means: agent steps and harness time in seconds. \emph{Cost} is \texttt{total\_cost\_usd} for the whole 152-task run; --- when the submission has no price.}
\label{tab:leaderboard}
\end{table*}



\paragraph{Leaderboard and extension.} Finished runs are exported to a public leaderboard with views for success, speed, and cost, broken down per domain and per interaction class. In the submission file the domain lists are the canonical tasks, and the interaction-class scores are the nine primary classes in \texttt{ui\_classes}. A new task is a JSON file; a new control implementation is one more registered value for a key, and every task in the domain can be cloned against it. The public leaderboard can be found on HuggingFace\footnote{\url{https://huggingface.co/spaces/ai-forever/WebPageBench}}. The codebase can be found in our repository\footnote{\url{https://github.com/ai-forever/WebPageBench}}.

\section{Evaluation}
\label{sec:evaluation}

\subsection{Protocol}
\label{sec:protocol}

A run fixes a harness, a model and a task set, and reports the \ems{}, Completion, steps, duration and cost. Those are the fields in the published submission, plus per-domain sections for its canonical task list and primary-class \texttt{ui\_classes}. Done\&pass is counted from the run logs. Over-claim, under-claim and harness-error counts are not in the submission. The runs ask whether event verification exposes failures hidden by the completion signal, how much performance varies across harnesses for one model, and how sensitive agents are to a control's implementation. Each submission's \ems{} covers all 152 tasks, variants included, but donors and variants are not scored separately. Static validation (\texttt{test\_verify\_bench}) checks the structure, routes and mounted profiles of all 152 task files. Decoding settings are the harness defaults; each pair is a single run. Model identifiers are those in Table~\ref{tab:leaderboard}.

\subsection{Results}
\label{sec:results}

Table~\ref{tab:leaderboard} reports the 24 public submissions. \ems{}, Completion, steps, duration and cost are submission fields; Done\&pass, the share of tasks the agent reported done and the log confirmed, is counted from the run logs. Within a model, bold marks the highest \ems{}, the shortest duration and the lowest cost.

\textbf{Completion and \ems{} diverge, and the sign of the gap depends on the harness.} Completion minus \ems{} reaches 41 points for \texttt{gpt-5.6-luna} with \texttt{openmanus} (1.00 vs.\ 0.59); its Done\&pass equals its \ems{}, so every verified success is hidden inside a completion signal that fires on every task. The same model with \texttt{openhands} is almost calibrated (0.638 vs.\ 0.625). On OpenHands the gap changes sign: Gemini's completion is 0.461 against an \ems{} of 0.572, and its Done\&pass of 0.454 means that 18 of its 87 verified successes ended without a completion signal. A protocol that equates the completion signal with success would inherit this harness-specific gap; the submission records the two rates separately.

\textbf{Harness ranking is model-dependent, and GUI still trails DOM.} \texttt{openmanus} is the best configuration for Gemini-3.8-flash (0.822) but not for GPT-5.6-luna, which peaks at \texttt{ouroboros-\allowbreak full-\allowbreak isolated} (0.796) and is worst on \texttt{openmanus} (0.592). Screenshot agents on the same 152-task \ems{} sit lower: Fara-1.5-9B at 0.638, OpenCUA-72B at 0.421, UI-TARS-1.5-7B at 0.336, EvoCUA below 0.10; Qwen3.8-27B scores 0.737--0.763 through DOM harnesses and 0.480 through screenshots.

\textbf{GUI agents also under-report the tasks they solve.} For the DOM harnesses Done\&pass is within one point of \ems{}: a solved task is a reported task. For the screenshot agents it is not: Fara-1.5-9B solves 0.638 but reports and solves only 0.493, Qwen3-VL 0.480 against 0.375, UI-TARS 0.336 against 0.289. Between one in nine and one in four verified successes of these agents ends without a completion signal.

\textbf{Cost and speed do not track \ems{}.} The most expensive run, \texttt{glm-5.2} with \texttt{openhands} (\$76.08), scores 0.500, while \texttt{gpt-5.6-luna} with \texttt{browser-use} scores 0.704 for \$3.23. Within one model the cheapest harness is not the best: for GPT-5.6-luna, \texttt{openmanus} costs least (\$3.04) and scores lowest. In each of the three model groups where it appears, OpenHands is the fastest harness per task (132--158~s) and at the same time the most expensive, because its per-task step count is high (131.7 steps for Gemini).

\textbf{Primary-class scores are the taxonomy the submission publishes.} On \texttt{gemini-3.8-flash} $\times$ \texttt{openmanus}, \texttt{ui\_classes} is BASKET 48/66, COUNTER 26/33, FILES 18/18, FAV 14/14, CARD 9/9, DATE 4/4, PAY 2/4, SELECT\_AC 2/2 and SELECT\_LIST 2/2 (Table~\ref{tab:taxonomy}). Classes with four or fewer tasks are diagnostics, not a ranking. The submission has no donor--variant split, so $\Delta_{h,w}$ is not reported.

\section{Conclusion}

\bench{} verifies web-agent tasks from the interface's own event log: no model decides success, and because the contract reads typed events and never the interface text, it does not depend on the language. Rendering one task through different implementations of one control makes sensitivity to interface form measurable. On the public 152-task leaderboard the completion--verification gap reaches 41 points, and the sign of the gap flips with the harness.

\section*{Limitations}
\label{sec:limitations}

\bench{} is a controlled set of mock sites, not a measurement of agents on production websites; anti-bot defenses, real payment rails and content drift are outside what it models. Events go to the backend, but application state (basket, favourites, login) lives in the browser's storage, so a scenario that spans two browser sessions is not expressible. The suite is Russian-only and does not test code-switching or transliteration beyond a single task family.

The published submissions do not split donor tasks from their variants, so this snapshot does not report $\Delta_{h,w}$. A variant changes DOM structure, action count and screenshot coordinates together, and a future split should not be read as proof that the agent misunderstood the task. The leaderboard covers 24 model--harness pairs. Each run is time-bounded. We accept that an agent may take a long time to answer; if it does not finish within the shared time limit, the task scores 0, and we treat that outcome as a normal part of the comparison. Event matching is not proof of understanding. Conditions are checked independently and without order, so a task that must distinguish two orderings of the same events cannot be expressed, and an agent could in principle emit the required events without understanding the task. Conditions that pin task-specific parameter values (item identifiers, dates, tariff names) make this unlikely, but we have not run an adversarial audit to confirm it.

\section*{Ethics Statement}

The environment contains no personal data: accounts, payment cards and passenger records are synthetic fixtures, and the payment flow is a mock that reaches no external system. Book catalog entries are replaced with synthetic authors and titles (\S\ref{sec:sites}). All 590 catalog photographs have a source URL and redistribution-compatible license in the release attribution workbook: CC~BY, CC~BY-SA, CC0 or Public Domain (\S\ref{sec:sites}). The software is released under the Apache License 2.0; third-party images retain their source licenses. The evaluation runs headless browsers against a local deployment and issues no traffic to third-party websites. 

We used generative AI tools while preparing this demonstration and paper (Cursor / Claude / GPT for drafting LaTeX and code review). All task files, event contracts, evaluation numbers and the wording of claims were checked by the authors; remaining errors are ours.

\section*{Acknowledgments}

We thank the contributors of the project: Maria Glushkova, Anna Kostikova, Elina Basyrova, Aleksandra Smolenkova,
Nadezhda Shervarly, Alexander Mazurin, Anastasia Filonova and Dmitri Baliev for
implementing the mock sites, the evaluation runner and the public leaderboard,
for authoring and validating the task suite, and for running the leaderboard
submissions. Compute for the 24 leaderboard pairs was provided by cloud.ru.

\begingroup\hbadness=2000 
\bibliography{custom}
\endgroup

\appendix

\section{Example Tasks}
\label{app:examples}

Tasks, interfaces and catalog data are Russian; the examples below are translated for readability, and the conditions are quoted verbatim. Table~\ref{tab:examples} shows one donor task per domain together with the variants generated from it.\footnote{Original prompts ship with the suite; the translation affects this paper only, not the evaluation.}

\begingroup\raggedright
\paragraph{Marketplace, one action.} \emph{Prompt:} ``add any catalog item to the cart.'' \emph{Condition:} one \texttt{basket\_\allowbreak add} event, no parameters required.

\paragraph{Hotels, four parameters.} \emph{Prompt:} ``show options in Hotels: Zurich, Switzerland, check-in 15.09, check-out 20.09, 2 guests.'' \emph{Conditions:} \texttt{bench\_\allowbreak hotel\_\allowbreak select\_\allowbreak city} (\texttt{cityId}, \texttt{cityName}, \texttt{countryName}), \texttt{bench\_\allowbreak hotel\_\allowbreak select\_\allowbreak start\_\allowbreak date}, \texttt{bench\_\allowbreak hotel\_\allowbreak select\_\allowbreak end\_\allowbreak date}, \texttt{bench\_\allowbreak hotel\_\allowbreak select\_\allowbreak guests} (\texttt{roomsCount=1}, \texttt{guestsCount=2}), and \texttt{state\_\allowbreak changed} to the search results.

\paragraph{Rail, seven conditions.} \emph{Prompt:} ``buy a Moscow--St.\ Petersburg ticket for 1 passenger on the economy tariff: add it to the cart and pay.'' \emph{Conditions:} \texttt{select\_\allowbreak city} (from), \texttt{select\_\allowbreak city} (to), \texttt{select\_\allowbreak date}, \texttt{select\_\allowbreak train}, \texttt{select\_\allowbreak tariff} (tariff name), \texttt{basket\_\allowbreak add}, \texttt{submit\_\allowbreak payment} (\texttt{result=success}).
\par\endgroup

\begin{table*}[t]
\centering
\small
\begin{tabular}{@{}>{\raggedright\arraybackslash}p{0.16\textwidth}>{\raggedright\arraybackslash}p{0.42\textwidth}>{\raggedright\arraybackslash}p{0.36\textwidth}@{}}
\toprule
\textbf{Donor task} & \textbf{Prompt (translated) and default controls} & \textbf{Released variants of this donor} \\
\midrule
Hotels \newline \texttt{hotel\_search\_} \texttt{scenario} &
``In Hotels, show options: Zurich, Switzerland, check-in 15.09.2026, check-out 20.09.2026, 2 guests.'' Defaults: split popup calendar, autocomplete city field, per-room guest popup. &
\texttt{date} $\to$ inline calendar, single-field range, typed string; \texttt{select\_city} $\to$ native \texttt{select}; \texttt{counter\_guests} $\to$ inline stepper, compact \texttt{select}, pill buttons; dark theme. (8 variants; four sibling hotel tasks with other cities and guest mixes carry the same profiles.) \\
\addlinespace
Rail \newline \texttt{rail\_book\_} \texttt{to\_cart} &
``In Trains, find a Moscow--St.\ Petersburg train on 04.10.2026, 1 passenger, Economy tariff, and add it to the cart.'' Defaults: grid popup calendar, typeahead station field. &
\texttt{date} $\to$ native \texttt{<input type="date">}, typed string; \texttt{select\_station} $\to$ native \texttt{select}; dark theme. (4 variants.) \\
\addlinespace
Books \newline \texttt{digital\_books\_} \texttt{named\_product\_} \texttt{basket} &
``In Books, find \emph{Quiet Amber in the Last Carriage} by Vera Rudneva (text edition) and add it to the cart.'' Default: standard search field. Author and title are synthetic (\S\ref{sec:sites}). &
\texttt{text\_search} $\to$ filled, outlined, pill, underlined; dark theme. (5 variants; the marketplace search tasks carry the same four search profiles.) \\
\addlinespace
Files \newline \texttt{files\_} \texttt{download\_pdf} &
``In Files, open \emph{Lab reports}, choose 2024 and download \texttt{lab-reports-2024.pdf}.'' Defaults: card layout, year \texttt{select}, outline buttons. &
one three-key profile: \texttt{collections} $\to$ list, \texttt{years} $\to$ buttons, \texttt{buttons} $\to$ icon; dark theme. (2 variants.) \\
\bottomrule
\end{tabular}
\caption{One donor per domain with its released variants. The prompt and the conditions of every variant are identical to the donor's; only the named control changes. Prompts are translated from Russian; identifiers are the task file names without the profile suffix.}
\label{tab:examples}
\end{table*}

\begin{table}[t]
\centering
\small
\resizebox{\linewidth}{!}{%
\begin{tabular}{@{}llr@{}}
\toprule
\textbf{Class} & \textbf{Control} & \textbf{Solved} \\
\midrule
\texttt{BASKET} & add / remove / quantity & 48/66 \\
\texttt{COUNTER} & stepper, guest count & 26/33 \\
\texttt{FILES} & collections, year, download & 18/18 \\
\texttt{FAV} & favourites toggle & 14/14 \\
\texttt{CARD} & card in a grid or carousel & 9/9 \\
\texttt{DATE} & date or range picker & 4/4 \\
\texttt{PAY} & payment form & 2/4 \\
\texttt{SELECT\_AC} & autocomplete list & 2/2 \\
\texttt{SELECT\_LIST} & click-to-select list & 2/2 \\
\bottomrule
\end{tabular}%
}
\caption{Primary classes in the public submissions. \emph{Solved} is passed/total from \texttt{ui\_classes} for \texttt{gemini-3.8-flash} $\times$ \texttt{openmanus} (best \ems{}). Every submission contains these nine classes and no others. Classes with four or fewer tasks are not a ranking. \texttt{NAV}, \texttt{RADIO}, \texttt{SEARCH} and \texttt{SEAT} are absent from \texttt{ui\_classes} and are omitted.}
\label{tab:taxonomy}
\end{table}

\section{Variant Catalog}
\label{app:variants}

Nine UI configuration keys (eight controls and the theme) carry 33 implementations: date (6: four on the hotel site, three on the rail site, with the typed-string input shared), text search (6), guest counter (4), collection layout (4), download button (4), year selector (3), city selector (2), station selector (2), theme (2). The 87 released variants instantiate 20 profiles over 25 donor tasks; 12 are dark-theme clones, two per domain, and five document-cabinet variants change layout, year selector and button style together because these three controls share one screen.

Counts in this paper are computed directly from the released task files: a task is a variant when its \texttt{test\_data} carries \texttt{ui\_variants} and \texttt{ui\_variant\_profile}, and canonical otherwise; per-domain counts group by \texttt{bench\_first\_domain}. Counted on the release tree: 65 canonical, 87 variants, 152 total; 19 event types and 521 conditions; 33 registered implementations across nine UI configuration keys (and 33 materialised profile JSON files).

\section{Interaction Classes}
\label{app:taxonomy}

Each task is annotated with the interaction classes it exercises, one of them designated as \emph{primary}. The taxonomy defines 19 classes. Table~\ref{tab:taxonomy} lists only the nine primary classes present in every public submission, with the solved/total score from the top pair. \texttt{NAV}, \texttt{RADIO}, \texttt{SEARCH} and \texttt{SEAT} occur in the task files but not in \texttt{ui\_classes}, so they are omitted. The other six (button, free text, checkbox, filter, login, low-level DOM activity) carry no released task.

\section{Related Benchmarks}
\label{app:related}

Table~\ref{tab:related} collects the comparison discussed in \S\ref{sec:related}. Rows are grouped by what decides success: a stored reference, a judge model, programmatic state plus a model, or a program alone. \bench{} belongs to the last group and differs within it, because a condition is matched against typed events emitted during the run, including intermediate steps and their parameters, rather than against the state left at the end. The two right-hand columns are where that comparison is sharpest. No other system in the table re-renders one fixed task through an alternative implementation of a control; WorkArena++ only changes colours and logos. The Agents column is not a count of models: it records how agents are attached, and where several can be run they usually arrive through one shared ecosystem, BrowserGym and AgentLab.

\begin{table*}[t]
\centering
\scriptsize
\setlength{\tabcolsep}{3pt}
\resizebox{\textwidth}{!}{%
\begin{tabular}{@{}lllcclll@{}}
\toprule
\textbf{Success decided by} & \textbf{Benchmark} & \textbf{Environment} & \textbf{Dom.} & \textbf{Tasks} & \textbf{Verification} & \textbf{Controlled UI variation} & \textbf{Agents} \\
\midrule
\multirow{2}{*}{\textbf{Reference match}}
 & Mind2Web \citep{deng2023mind2web} & offline, branded & 31 & 2{,}350 & trajectory match & --- & --- \\
 & AssistantBench \citep{yoran2024assistantbench} & live, branded & 258 & 214 & answer match & --- & BrowserGym \\
\midrule
\multirow{4}{*}{\textbf{Judge model}}
 & WebVoyager \citep{he2024webvoyager} & live, branded & 15 & 643 & GPT-4V judge & --- & --- \\
 & InSTA \citep{trabucco2025insta} & live & ${\sim}150$K & 146{,}441 & LLM judge & --- & Gym env \\
 & BookingArena \citep{logeswaran2026bookingarena} & live, branded & 20 & 120 & VLM constraint judge & --- & --- \\
 & GTA \citep{huang2026gta} & live, crawled & 50+ & 5{,}600 & answer + path replay & --- & --- \\
\midrule
\multirow{4}{*}{\textbf{Hybrid}}
 & WebArena \citep{zhou2023webarena} & hosted, de-branded & 4 (+3) & 812 & state + fuzzy LLM & --- & BrowserGym \\
 & VisualWebArena \citep{koh2024visualwebarena} & hosted, de-branded & 3 & 910 & + VQA / SSIM judge & --- & BrowserGym \\
 & REAL \citep{garg2025real} & hosted, renamed & 11 & 112 & state diff + LLM rubric & --- & 2, pluggable \\
 & TheAgentCompany \citep{xu2024theagentcompany} & hosted, branded & 4 & 175 & scripts + LLM (29\% tasks) & --- & 1 agent \\
\midrule
\multirow{5}{*}{\textbf{Deterministic}}
 & WebShop \citep{yao2022webshop} & simulated, de-branded & 1 & 12{,}087 & attribute reward & --- & Gym env \\
 & WorkArena++ \citep{boisvert2024workarena} & live, branded & 1 & 682 & backend final state & 10 brands (theme only) & AgentLab \\
 & AutoWebWorld \citep{wu2026autowebworld} & generated & 29 & 11{,}663 traj. & state-machine replay & --- & --- \\
 & WebForge \citep{yuan2026webforge} & generated & 7 & 934 & final-state compare & --- & --- \\
 & \textbf{\bench{} (ours)} & \textbf{hosted mocks, de-branded} & \textbf{6} & \textbf{65 + 87 var.} & \textbf{event-level match} & \textbf{8 controls + theme, 33 impl.} & \textbf{6 browser/DOM + 5 GUI eval.} \\
\midrule
\emph{No environment} & BrowserGym / AgentLab \citep{chezelles2025browsergym} & harness ecosystem & --- & --- & --- & --- & ${\sim}10$ agents \\
\bottomrule
\end{tabular}}
\caption{Web-agent benchmarks grouped by what decides success (first column): a reference that never inspects the environment, a judge model, programmatic state plus a model, or a program alone. \emph{Environment}: live, offline, self-hosted, simulated or generated sites and their branding; \emph{Dom.}: distinct domains or sites; \emph{Agents}: how agents are connected (descriptive, not a count). Task counts as reported by each paper. Among the systems surveyed, none re-renders a fixed task through an alternative implementation of a control (WorkArena++ changes colours and logos only), and multi-agent support comes through one shared ecosystem.}
\label{tab:related}
\end{table*}
\end{document}